%% file: icml2026.tex
\documentclass{article}

\usepackage{microtype}
\usepackage{graphicx}
\usepackage{subcaption}
\usepackage{booktabs} 

\usepackage{hyperref}

\usepackage[accepted]{icml2026}

\usepackage{amsmath}
\usepackage{amssymb}
\usepackage{mathtools}
\usepackage{amsthm}

\usepackage[capitalize,noabbrev]{cleveref}

\theoremstyle{plain}

\theoremstyle{definition}

\theoremstyle{remark}

\usepackage[textsize=tiny]{todonotes}

\usepackage{colortbl}
\usepackage{multirow}
\usepackage{array}
\usepackage{rotating}
\usepackage{caption}
\usepackage{makecell}
\definecolor{familybg}{gray}{0.82}
\definecolor{rowalt}{gray}{0.95}
\newcolumntype{L}[1]{>{\raggedright\arraybackslash}p{#1}}
\newcolumntype{C}[1]{>{\centering\arraybackslash}p{#1}}
\newcolumntype{R}[1]{>{\raggedleft\arraybackslash}p{#1}}
\newcommand{\familyrow}[1]{%
  \multicolumn{5}{>{\columncolor{familybg}}l}{\textbf{#1}}\\
}
\DeclareMathOperator*{\argmin}{arg\,min}

\begin{document}

\twocolumn[
  \icmltitle{Language Models Represent and Transform Concepts with Shared Geometry}



  \icmlsetsymbol{equal}{*}


  \begin{icmlauthorlist}
    \icmlauthor{Zhimin Hu}{GT}
    \icmlauthor{Lanhao Niu}{Edin}
    \icmlauthor{Sashank Varma}{GT}
  \end{icmlauthorlist}

  \icmlaffiliation{GT}{Georgia Institute of Technology}
  \icmlaffiliation{Edin}{University of Edinburgh}

  \icmlcorrespondingauthor{Zhimin Hu}{zhu41@gatech.edu}
  \icmlcorrespondingauthor{Sashank Varma}{varma@gatech.edu}

  \icmlkeywords{Geometry, Representation, LLM, Concept, Context}

  \vskip 0.3in
]



\printAffiliationsAndNotice{}  

\begin{abstract}
How concepts are represented in neural networks is a fundamental question in machine learning. The dominant view treats concept representations as stationary geometric objects. Yet concepts appear in context, and context transforms them. Drawing from neural population geometry, we formalize concept representations as point-cloud manifolds and contextual transformations as vector fields, and instantiate this framework in large language models. Across six model families of varying scales, we find that context moves each concept differently. The variance in these displacements is semantically organized, correlating with lexical concreteness and density. Importantly, both the concepts being transformed and this variance structure are shared across models: displacement structure transported from one model predicts held-out displacements in others significantly above chance. Together, these findings show that models share a common geometry not only in how concepts are represented, but more importantly in how context transforms them, a structure with richer organization than prior work has recognized.
\end{abstract}

\input{main}

\nocite{langley00}

\bibliography{reference}
\bibliographystyle{icml2026}

\newpage
\appendix
\onecolumn

\input{appendix}

\end{document}

%% file: main.tex
\section{Introduction}

How does a neural network represent concepts? The dominant view in machine learning treats concepts as stable geometric objects, such as vectors in activation space that persist across contexts \citep{park2024linear, huh2024position}. This answer has been productive. It grounds methods \citep{arditi2024refusal} to steer model behavior via single vectors and motivates alignment research \citep{jhaharnessing}. It also fits the modern cognitive science view of concepts as static vectors \citep{piantadosi2024concepts}. However, this is a stationary answer to a question that is fundamentally dynamic \citep{truman2024flexible}. A concept does not appear in isolation; it appears in diverse contexts, and contexts move it. 

\begin{figure*}
  \centering
  \includegraphics[width=0.95\textwidth]{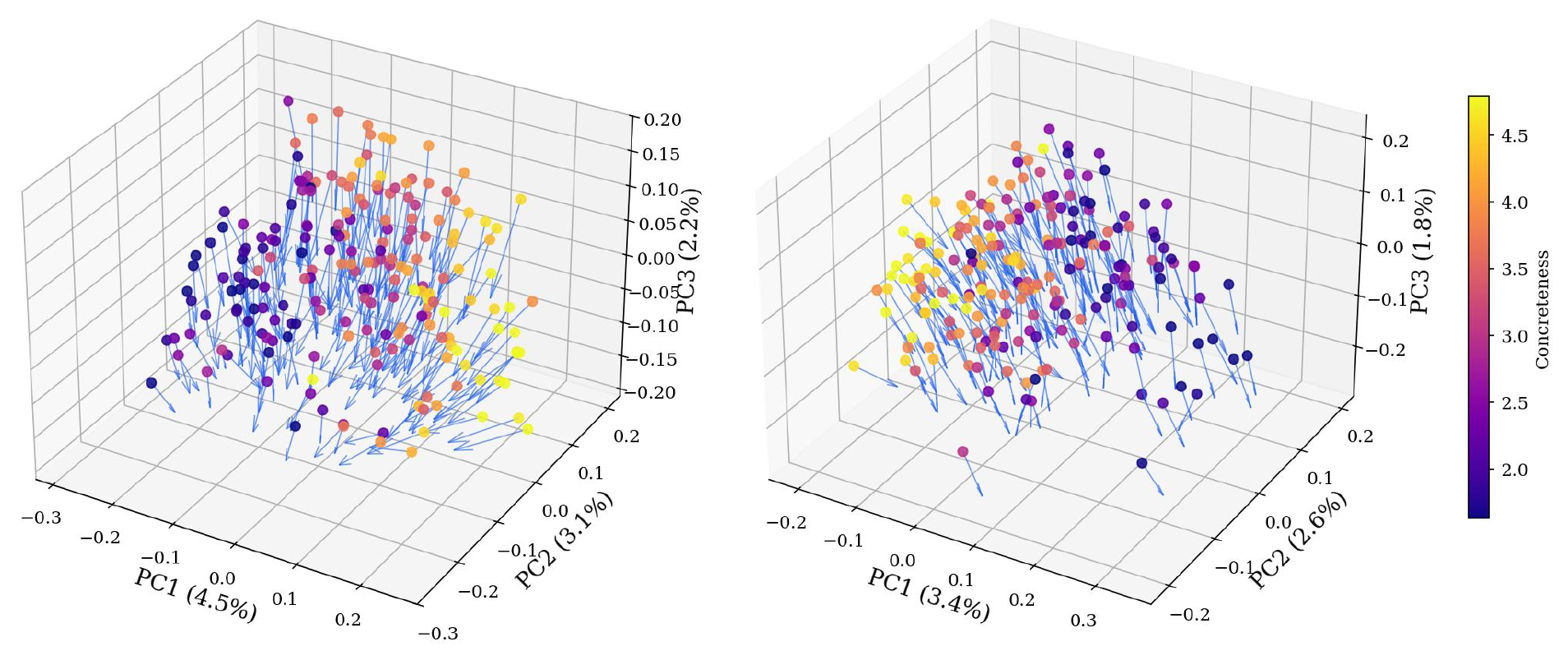} 
  \caption{Visualization of the contextual transformation field projected onto the top three principal components (left: Ministral3-8B; right: Qwen2.5-7B). Each point represents a concept word in its neutral context; the arrow emanating from each point is the displacement vector, projected into the same subspace. Point color encodes lexical concreteness. It illustrates that no single direction dominates the field. Arrows vary in both magnitude and direction across the vocabulary.}
  \label{fig:field}
\end{figure*}

In this paper, we examine this stationary view directly and find that it does not hold as prior work implies. Context does not perturb concepts around a stable vector. Rather, each concept moves in its own direction and by its own magnitude, and no single vector captures this variation (see Figure~\ref{fig:field}). Yet these variances are not arbitrary. They are systematically structured, correlating with lexical concreteness and density. Despite this heterogeneity, the relational geometry is shared across models for both the concepts and the displacements that transform them: how concepts sit and move relative to one another is consistent even when their absolute positions and displacements are not.

We verify the above findings by showing that a relational procedure transporting displacement structure from one model predicts held-out displacements in another significantly above baseline, and that distorting this relational structure degrades prediction. We further show that the sharing of conceptual representations is not a surface artifact - it cannot be reduced to co-occurrence statistics, and scrambling the context collapses the alignment. To summarize, what is consistent across models includes not only where concepts are, but the geometry of how contexts transform them, a structure that has not, to our knowledge, been systematically characterized before.

These results matter at two scales. At the level of representation engineering, we provide a geometric account of single-vector steering as a useful approximation that nonetheless discards a structured residual that is semantically meaningful. More broadly, our results speak to the longstanding puzzle in cognitive science of how concepts can be simultaneously stable and flexible \citep{casasanto2015all}. We provide a potential answer. Concepts are stable in the sense that their relation to one another remain consistent. They are flexible in the sense that context moves each concept in a semantically different way.

\section{Preliminaries}
\label{sec:preliminaries}

\subsection{Concepts as Point-Cloud Manifolds}

In neural population geometry, a stimulus category corresponds
to a \textit{point-cloud manifold} in neural state space, where
a set of population responses arises from identity-preserving
variability in the input \citep{chung2021neural}. Despite the implied underlying geometry, these structures manifest empirically as point clouds due to sparse sampling.

We adopt an analogous construction for concept representations
in language models. Let $\mathcal{M}^{(m)} \subseteq
\mathbb{R}^{d_m}$ denote the representation space of model $m$,
where $d_m$ is the hidden dimension at a given layer. Formally, let $f^{(m)}$ denote the forward pass of model $m$. For a concept word $w$ appearing in context $c$, the representation $r^{(m)}(w, c)$ is the contextualized embedding of $w$ under $f^{(m)}$ extracted at the hidden state 
corresponding to $w$'s token position. For multi-token words, $r^{(m)}(w, c)$ is the mean of the subword tokens. The set of such instantiations,
\begin{equation}
    \mathcal{X}_w^{(m)} = \{r^{(m)}(w, c_i)\}_{i=1}^{N}
    \subset \mathcal{M}^{(m)},
\end{equation}
forms a point cloud that we treat as an empirical sample from an underlying distribution $\mathcal{P}_w^{(m)}$ over $\mathcal{M}^{(m)}$. The parametric form of $\mathcal{P}_w^{(m)}$ is not assumed; thus, we characterize its geometry non-parametrically through pairwise kernel relations within $\mathcal{X}_w^{(m)}$ through kernel matrices. 

\subsection{Two Axes of Representational Geometry}

Following the formalization above, we characterize conceptual representation along two complementary axes, each reflecting a different property of how a concept is distributed - the variation \emph{within} a single concept across contexts, and the relations \emph{between} concepts within a shared context.

\paragraph{Within-Concept Relation.}

The within-concept relation concerns the shape of its
point cloud under distributional context variation. Given a context set $\mathcal{C}$, the point cloud
$\mathcal{X}_w^{(m,k)} = \{r^{(m)}(w, c_i)\}_{c_i \in
\mathcal{C}_k}$ yields a kernel matrix used to measure the contextual variability of the same concept.:
\begin{equation}
    K_w^{(m)}[i,j] = k\bigl(r^{(m)}(w, c_i),\,
    r^{(m)}(w, c_j)\bigr), \quad c_i, c_j \in \mathcal{C}.
\end{equation}

\paragraph{Between-Concept Relation.}

The between-concept relation concerns how concepts are positioned relative to one another when embedded in the same context. For a given context $\tau$, we form the kernel matrix used to encode the pairwise relation of concepts:
\begin{equation}
    K_\tau^{(m)}[i,j] = k\bigl(r^{(m)}(w_i, \tau),\,
    r^{(m)}(w_j, \tau)\bigr), \quad w_i, w_j \in \mathcal{W},
\end{equation}

\subsection{Contextual Transformations as Vector Fields}
Let $\tau_0 \in \mathcal{T}$ denote the source context.
The \textit{contextual displacement} of concept $w$ under
target context $\tau$ in model $m$ is:
\begin{equation}
\label{eq:transformation}
    \phi^{(m)}(w, \tau) = r^{(m)}(w, \tau) -
    r^{(m)}(w, \tau_0) \in \mathcal{M}^{(m)},
\end{equation}
Across the concept vocabulary $\mathcal{W}$, the collection of displacements defines a vector field:
\begin{equation}
    \Phi_\tau^{(m)} : \mathcal{W} \rightarrow \mathcal{M}^{(m)},
    \quad w \mapsto \phi^{(m)}(w, \tau).
\end{equation}
This vector field thus represents the \textit{contextual transformation} between $r^{(m)}(w, \tau)$ and $r^{(m)}(w, \tau_0)$. As usual, we characterize it by the kernel matrix:
\begin{equation}
    K^{\Phi_\tau^{(m)}}[i,j] = k\bigl(\phi^{(m)}(w_i, \tau),\,
    \phi^{(m)}(w_j, \tau)\bigr),
\end{equation}

\section{Measurement}
\label{sec:measurement}

Here, we specify how we compare the above-mentioned quantities across models and how we ensure the comparisons are not inflated by statistical artifacts.

\paragraph{Kernel Alignment.}
We denote $K$ and $L$ as the generic forms of kernel matrices from two models. Each takes the form $R R^\top$, where $R$ is the matrix of representations defined in Section~\ref{sec:preliminaries}. To capture the geometry that is invariant to the particular basis, rotation, or scaling of each model, we use Centered Kernel Alignment (CKA) 
\citep{kornblith2019similarity}. We defined our CKA via the unbiased 
HSIC estimator \citep{song2012feature} to avoid finite-sample inflation:
\begin{equation}
\begin{split}
    \widehat{\text{HSIC}}(K, L) =\;&
    \frac{1}{n(n-3)} \Bigg[
        \operatorname{tr}\tilde{K}\tilde{L}
        + \frac{\mathbf{1}^\top \tilde{K}
        \mathbf{1}\mathbf{1}^\top \tilde{L} \mathbf{1}}
        {(n-1)(n-2)} \\
        &- \frac{2}{n-2}\,
        \mathbf{1}^\top \tilde{K} \tilde{L} \mathbf{1}
    \Bigg],
\end{split}
\end{equation}
where $\tilde{K}$ and $\tilde{L}$ are the kernel matrices with zero diagonal entries. CKA is then defined as:
\begin{equation}
    \text{CKA}(K, L) = \frac{\widehat{\text{HSIC}}(K, L)}
    {\sqrt{\widehat{\text{HSIC}}(K, K) \cdot
    \widehat{\text{HSIC}}(L, L)}} \in [0, 1].
\end{equation}

\paragraph{Grassmann Distance.}
CKA captures the relational structure of representations. To characterize whether contextual transformations span the same directions in their respective representation spaces, we employ the Grassmann distance \citep{hamm2008grassmann}. Recall that $\Phi_\tau^{(m)} \in \mathbb{R}^{|\mathcal{W}| \times d_m}$
denotes the contextual transformation field for model $m$. The
principal subspace of $\Phi_\tau^{(m)}$ is obtained via
truncated SVD:
\begin{equation}
    \Phi_\tau^{(m)} = U^{(m)} \Sigma^{(m)} (V^{(m)})^\top,
    \quad U^{(m)} \in \mathbb{R}^{|\mathcal{W}| \times p},
\end{equation}
where $p$ is the dimension of the subspace, and we estimate it from the intrinsic dimension of $\Phi_\tau^{(m)}$ \citep{facco2017estimating}. The column space $\mathcal{S}^{(m)} = \text{col}(U^{(m)})$ is a point on the
Grassmann manifold $\text{Gr}(p, |\mathcal{W}|)$. The
principal angles $\{\theta_i\}_{i=1}^p$ between
$\mathcal{S}^{(m)}$ and $\mathcal{S}^{(m')}$ are defined via $\cos\theta_i = \sigma_i\bigl((U^{(m)})^\top U^{(m')}\bigr)$, where $\sigma_i$ denotes the $i$-th singular value. The
Grassmann distance is then:
\begin{equation}
    d_{\text{Gr}}\bigl(\mathcal{S}^{(m)},
    \mathcal{S}^{(m')}\bigr) = \left(\sum_{i=1}^p
    \theta_i^2\right)^{1/2}.
\end{equation}

\paragraph{Permutation Calibration.}

Let $\rho_\text{obs} = \text{sim}(R^{(m)}, R^{(m')})$ denote the observed alignment between two matrices, where $\text{sim()}$ can be either CKA or Grassmann distance. We generate $K = 200$ permutations $\{\pi_k\}_{k=1}^K$ and evaluate the null distribution $\{\rho_k\}_{k=1}^K$. Following \citep{groger2026revisiting}, the critical threshold is taken as the $(1-\alpha)$-quantile of the combined distribution $\gamma = Q_{1-\alpha}\bigl(\{\rho_\text{obs}\} \cup \{\rho_k\}_{k=1}^K\bigr)$, with add-one p-value $p = (1 + \sum_k \mathbf{1}[\rho_k \geq \rho_\text{obs}]) / (K+1)$. The calibrated score is then:
\begin{equation}
    \rho_\text{cal} = \frac{(\rho_\text{obs} - \gamma)_+}
    {\rho_\text{max} - \gamma},
\end{equation}
where $(\cdot)_+ = \max(0, \cdot)$ and $\rho_\text{max}$ is the maximum attainable similarity. For Grassmann distance, calibration yields the threshold $\lambda_d$ directly; the reported statistic is the margin $\lambda_d-d_{\mathrm{Gr}}$.

\section{Material}

\subsection{Datasets and Stimuli}

\paragraph{Within-Concept Relation.}
We operationalize concepts along artificial vs.\ natural and concrete vs.\ abstract dimensions. We sample concepts uniformly from human concreteness ratings \citep{brysbaert2014concreteness} and map them with artificial and natural concepts sampled from WordNet \citep{miller1995wordnet}. We construct a vocabulary of 40 concepts with approximately uniform concreteness distribution (see Appendix~\ref{app:vocabulary} for details). For each concept $w$, we sample 100 positionally balanced sentence chunks from Wikipedia, excluding sentence-initial occurrences to control positional bias in decoder-only models.

\paragraph{Between-Concept Relation.}
We sample approximately 300 words from the Google News Word2Vec \citep{mikolov2013efficient}, stratified into three subsets of 100 words by lexical density, estimated as the mean pairwise cosine similarity. This partition mirrors the data size of the above section and allows downstream analyses to assess whether alignment in relational structure varies with lexical density.

Since naturalistic contexts vary idiosyncratically across words and cannot be held constant across the vocabulary, we design controlled prompt templates grounded in theories of conceptual representation \citep{rosch1978cognition, barsalou1999perceptual, barsalou2003situated, osgood1957measurement, murphy2004big}, spanning categorization, perception, situation, affect, and world knowledge. These prompt-based templates have been shown to elicit human-like conceptual representations from language models \citep{hu2024failures, xu2025revealing, hu2026representational}. Each template places $w$ at the penultimate position. The neutral template reads ``Thinking of the following concept: $w$.''\ and serves as the reference context $\tau_0$ in Equation~(\ref{eq:transformation}). The remaining five templates foreground: categorization (``Thinking of the conceptual category of $w$.''), perception (``Thinking of the sensory and perceptual properties of $w$.''), situation (``Thinking of a typical situation involving $w$.''), affection (``Thinking of the feelings and associations evoked by $w$.''), and knowledge (``Thinking of general world knowledge about $w$.'').

\paragraph{Contextual Transformation.}
\label{subsec:template}
To probe the $\Phi_\tau^{(m)}$, we require a large vocabulary to support analysis of how lexical properties relate to transformation structure. We therefore draw from two complementary sources: the Google News word2vec vocabulary \citep{mikolov2013efficient}, which captures distributional co-occurrence structure and provides a grounded basis for lexical density; and Brysbaert's concreteness rating \citep{brysbaert2014concreteness}, which provides an independent semantic property orthogonal to distributional structure. Words are filtered to exclude informal and non-standard forms, yielding approximately 1000 words \citep{bird2009natural}. Lexical density and concreteness are stratified into bins and balanced to remove their linear dependence; the resulting vocabulary achieves a Pearson correlation of $r = -0.018$ between the two variables, confirming effective decorrelation.

\subsection{Language Models}
We conduct our measurement over 23 publicly available base pretrained models spanning six families, ranging from 0.5B to 32B parameters. We restrict ourselves to base models to avoid representational shifts introduced by instruction tuning. To enable within-family scaling analysis, we require dense coverage across the full scale range. Thus, the Qwen2.5 family serves as the backbone model since it satisfies this requirement with six checkpoints \cite{qwen2025qwen25}. The remaining models span Llama (Llama2-7B/13B, Llama3/3.1-8B) \cite{touvron2023llama, grattafiori2024llama}, Mistral (Ministral3-3B/8B/14B, Mistral-Small-24B) \cite{liu2026ministral}, Gemma2 (2B, 9B) \cite{team2024gemma}, and DeepSeek-LLM (7B) \cite{bi2024deepseek}. The Pythia suite (0.4B--12B) \cite{biderman2023pythia} is included as an older-generation reference; these models are excluded from capability-based analyses due to the lack of MMLU scores. Full details on model selection are in Appendix~\ref{app:models}.

\begin{table*}[h]
\centering
\caption{Spearman correlation between word properties and transformation field metrics across models and layer depths. Significance: $^{*}p<0.05$, $^{**}p<0.01$, $^{***}p<0.001$.}
\label{tab:property_correlation}
\resizebox{0.92\textwidth}{!}{%
\begin{tabular}{ll ccc ccc}
\toprule
& & \multicolumn{3}{c}{Magnitude $\tilde{r}$} & \multicolumn{3}{c}{Directional deviation $\sigma$} \\
\cmidrule(lr){3-5} \cmidrule(lr){6-8}
Property & Model & 25\% & 50\% & 75\% & 25\% & 50\% & 75\% \\
\midrule
\multirow{4}{*}{Concreteness}
& Qwen2.5-7B     & $-0.09^{**}$  & $-0.04$       & $0.22^{***}$  & $-0.41^{***}$ & $-0.24^{***}$ & $-0.27^{***}$ \\
& Llama3.1-8B    & $-0.07^{*}$   & $0.07^{*}$    & $0.10^{**}$   & $-0.36^{***}$ & $-0.32^{***}$ & $-0.25^{***}$ \\
& Ministral3-8B  & $-0.13^{***}$ & $-0.07^{*}$   & $0.09^{**}$   & $-0.32^{***}$ & $-0.25^{***}$ & $-0.21^{***}$ \\
& Gemma2-9b      & $-0.11^{***}$ & $0.13^{***}$  & $0.07^{*}$    & $-0.22^{***}$ & $-0.23^{***}$ & $-0.28^{***}$ \\
\midrule
\multirow{4}{*}{Density}
& Qwen2.5-7B     & $-0.32^{***}$ & $-0.32^{***}$ & $-0.27^{***}$ & $-0.11^{***}$ & $-0.19^{***}$ & $-0.16^{***}$ \\
& Llama3.1-8B    & $-0.28^{***}$ & $-0.28^{***}$ & $-0.22^{***}$ & $-0.14^{***}$ & $-0.15^{***}$ & $-0.10^{**}$  \\
& Ministral3-8B  & $-0.30^{***}$ & $-0.28^{***}$ & $-0.26^{***}$ & $-0.23^{***}$ & $-0.18^{***}$ & $-0.16^{***}$ \\
& Gemma2-9b      & $-0.28^{***}$ & $-0.27^{***}$ & $-0.31^{***}$ & $-0.20^{***}$ & $-0.22^{***}$ & $-0.16^{***}$ \\
\bottomrule
\end{tabular}%
}
\end{table*}

\section{Results}
We present results in order of the central argument. We first characterize the structure of contextual transformation fields and establish that their variance is semantically organized and shared across models. We then demonstrate that the representations being transformed are themselves shared across models for both within and between-concept relations, and that this sharing is not a statistical artifact.

\subsection{Shared Geometry of Contextual Transformations}
\label{subsec:transform}

Since cross-model alignment in both within and between-concept relations is stable at matched normalized layer depths (Section~\ref{subsec:geometry}), and middle-to-late layers are known to exhibit stable semantic structure and brain alignment \citep{caucheteux2022brains}, we report transformation field analyses at three representative depths: 25\%, 50\%, and 75\% of each model's total depth. Primary conclusions are drawn from the 50\% and 75\% depths, where semantic organization is most pronounced.

\paragraph{The transformation field is not uniform, and the variance is semantically organized.} To characterize the uniformity of $\Phi_\tau^{(m)}$ and assess whether its variance structure is semantically organized, we quantify field dispersion via the leading variance ratio $\rho_1$, the variance explained by first principal component \citep{abdi2010principal}, and the global spherical variance $V_{\text{global}}$, the mean squared geodesic distance from each unit-normalized displacement to the spherical 
Fréchet mean (see Appendix~\ref{app:nonlinear} for details). Across all models and depths, $\rho_1 \in [0.073, 0.146]$ and $V_{\text{global}} \in [0.43, 0.75]$ rad$^2$ (cf.\ $\pi^2/4 \approx 2.47$ 
for a uniform distribution), confirming that no single direction dominates the field.

We next ask whether the variance structure is semantically organized. For each word $w$, we compute two per-word statistics from $\Phi_\tau^{(m)}$: the normalized displacement magnitude $\tilde{r}_w=\left\|\phi^{(m)}(w, \tau)\right\| /\left\|r^{(m)}\left(w, \tau_0\right)\right\|$, and the directional deviation $\sigma_w$, defined as the geodesic distance from $\hat{\phi}^{(m)}(w, \tau)$ to the field's Fr\'{e}chet mean. We correlate each measure against concreteness ratings \citep{brysbaert2014concreteness} and lexical density estimated as mean pairwise cosine similarity in Word2Vec. We separate the data by bins based on concreteness ratings and lexical density and obtain the within-bin version of $\Phi_\tau^{(m)}$ and $\sigma_w$ to capture the correlation as semantic groups.

\begin{figure*}[h]
  \centering
  \includegraphics[width=0.95\textwidth]{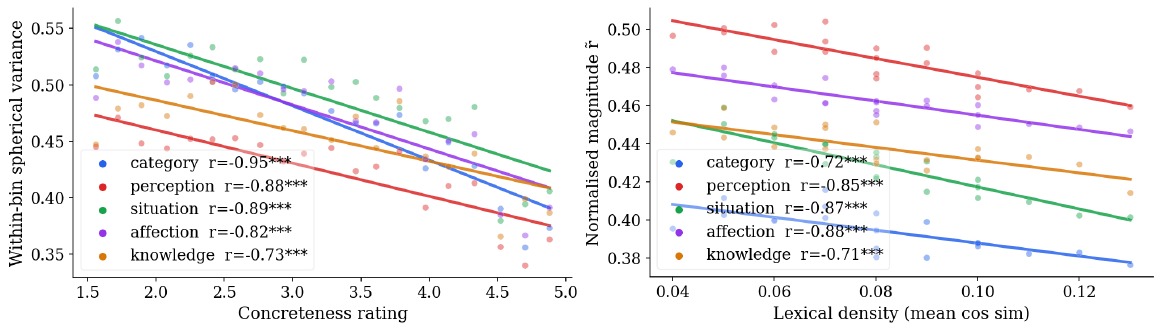} 
  \caption{Spearman correlation between word properties and within bin transformation field metrics at layer depth 75\% of Qwen2.5-7B. Left: within bin $\sigma_w$ as a function of concreteness rating. Right: within bin $\tilde{r}$ as a function of lexical density. Regression lines are shown for each context.}
  \label{fig:variance}
\end{figure*}

As shown in Table~\ref{tab:property_correlation}, lexical density correlates negatively with displacement magnitude ($\rho \in[-0.22,-0.32], p<0.001$) across all entries, indicating that concepts in denser semantic neighborhoods are displaced less by context. Concreteness correlates negatively with directional deviation ($\rho \in[-0.21,-0.41], p<0.001$) across all entries, indicating that more concrete concepts exhibit less directional spread across contexts. Figure~\ref{fig:variance} confirms that the within-bin version of $\Phi_\tau^{(m)}$ and $\sigma_w$ and  are graded. The linear relationships hold across all five semantic context dimensions. Together, these results suggest a geometric account of conceptual flexibility. Abstract concepts, lacking strong perceptual grounding, are displaced in more variable directions across semantic framings, while dense concepts resist large absolute displacements. The variance in the transformation field is a semantically organized property of the lexicon, not arbitrary noise.

\paragraph{The alignment of the transformation field scales with model capability.} We next ask whether the geometry of the transformation field is shared across models of different scales and capacities, and whether alignment tracks model size or functional competence. We designate Qwen2.5-32B as the reference model and compare its transformation field against all other models as it achieves the highest MMLU score in our set, providing the most capable reference point for measuring convergence.

\begin{figure*}[h]
  \centering
  \includegraphics[width=0.95\textwidth]{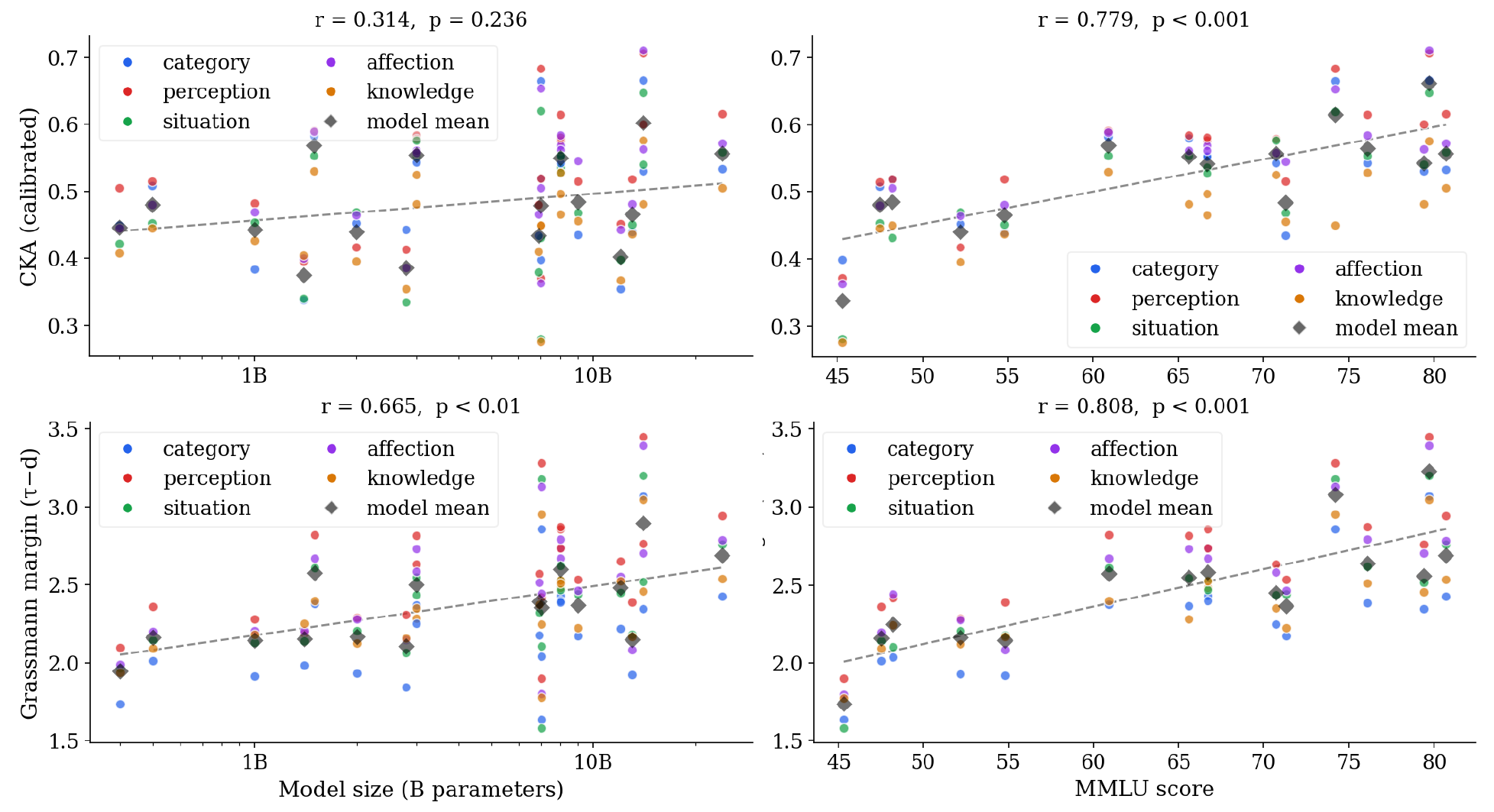} 
  \caption{Alignment of the transformation field between Qwen2.5-32B and other models, shown as a function of model size (left column) and MMLU score (right column). Top row: CKA. Bottom row: Grassmann margin ($\lambda_d - d$). Filled diamonds indicate per-model means averaged across contexts.}
  \label{fig:scaling}
\end{figure*}

As shown in Figure~\ref{fig:scaling}, model capability as measured by MMLU \citep{hendrycks2020measuring} is a substantially stronger predictor of alignment than model size: CKA versus MMLU yields $r=0.779$, $p<0.001$; Grassmann margin versus MMLU yields $r=0.808$, $p<0.001$. The corresponding size-based correlations are weaker ($r=0.314$, $p=0.236$ for CKA; $r=0.665$, $p<0.01$ for Grassmann), with the CKA trend failing to reach significance. This dissociation indicates that representational alignment in the transformation field tracks functional competence more closely than parameter count, though we cannot rule out that capability and training data quality are confounded at scale. It suggests that the shared geometry of transformation reflects the learned semantic organization of models

\paragraph{The relational structure of the transformation field is transferable across models.} The preceding results establish that transformation fields are semantically organized and that they are aligned across models. We now directly validate the alignment by using the relational structure of the transformation field in one model to predict the held-out displacement in different models. 

We address this through a relational transport procedure. 
For source model $m$ and target model $m'$, we partition 
$\mathcal{W}$ into train and test splits (80/20). For each 
test word $w_{\text{test}}$, we predict its displacement 
in $m'$ as a cosine-similarity-weighted combination of 
training displacements in $m'$, where weights are derived 
entirely from the relational structure of the source model:
\begin{equation}
    \hat{\phi}^{(m')}(w_{\text{test}}, \tau) = 
    \frac{\sum_{w_j \in \mathcal{W}_{\text{train}}} 
    s_j^{(m)}(w_{\text{test}}) \cdot \phi^{(m')}(w_j, \tau)}
    {\sum_{w_j \in \mathcal{W}_{\text{train}}} 
    |s_j^{(m)}(w_{\text{test}})|},
\end{equation}
where $s_j^{(m)}(w_{\text{test}})$ is the cosine similarity between displacements in the source model. We evaluate predictions on two metrics. The first is magnitude rank correlation $\rho$, the Spearman correlation 
between $\|\hat{\phi}^{(m')}\|$ and $\|\phi^{(m')}\|$ over test words. The second is residual cosine similarity, defined as the mean cosine between predicted and true displacements after subtracting the training-split mean, which isolates word-specific directional deviation from the field mean. We compare this relational transport against a random permutation baseline, which uses the same weighted combination but with randomly shuffled source weights, and a random Gaussian baseline, which replaces the predicted displacement with a Gaussian noise vector matched to the magnitude of the training set.

\begin{figure*}
  \centering
  \includegraphics[width=0.95\textwidth]{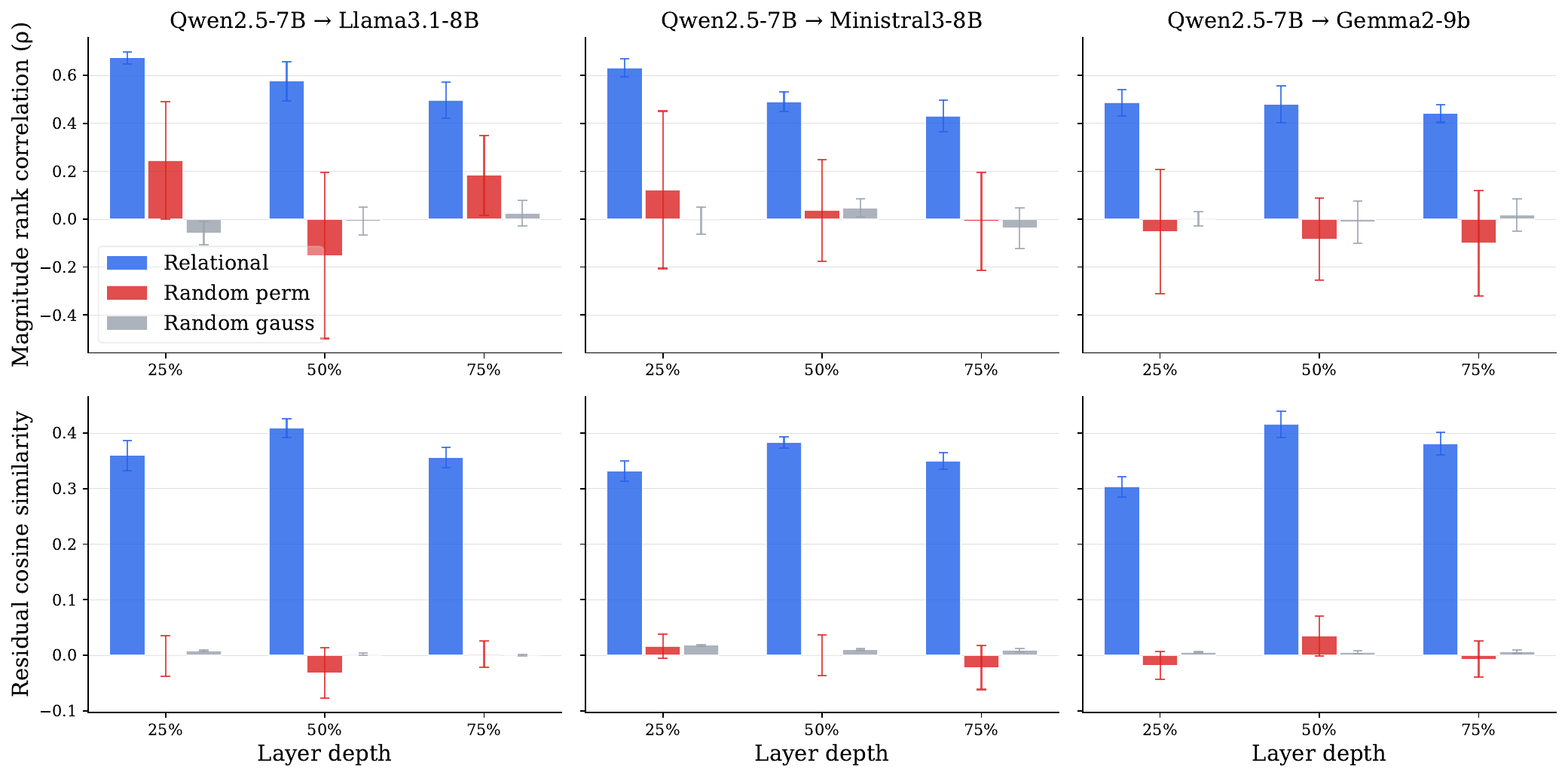} 
  \caption{Relational transport of transformation structure from Qwen2.5-7B to three target models. Top row: rank correlation of magnitude between predicted and ground-truth displacement on held-out test words. Bottom row: cosine similarity between the residual of predicted and ground-truth displacement directions from the mean field. Blue bars show the relational transport procedure; red bars show a random permutation ablation; gray bars show a random Gaussian baseline. Error bars denote standard deviation across contexts.}
  \label{fig:transport}
\end{figure*}

In Figure~\ref{fig:transport}, we report two evaluation metrics, the magnitude rank correlation $\rho$ and residual cosine similarity, across three target models (Llama3.1-8B, Ministral3-8B, Gemma2-9B) and three layer depths. The relational method substantially and consistently outperforms both baselines on both metrics. Magnitude rank correlations for the relational method are positive and well above zero across all conditions, while both ablation baselines cluster near zero, with the random permutation baseline occasionally negative. Residual cosine similarity follows the same pattern: the relational method yields positive similarity between predicted and true displacements, whereas both baselines remain near zero. Both effects are robust across layer depths. This directly validates our central claim that the geometry of how context transforms concepts is a stable structure of concept space, not a statistical artifact of shared vocabulary or global field magnitude.

\subsection{Shared Geometry of Concept Spaces}
\label{subsec:geometry}

\begin{figure*}
  \centering
  \includegraphics[width=0.95\textwidth]{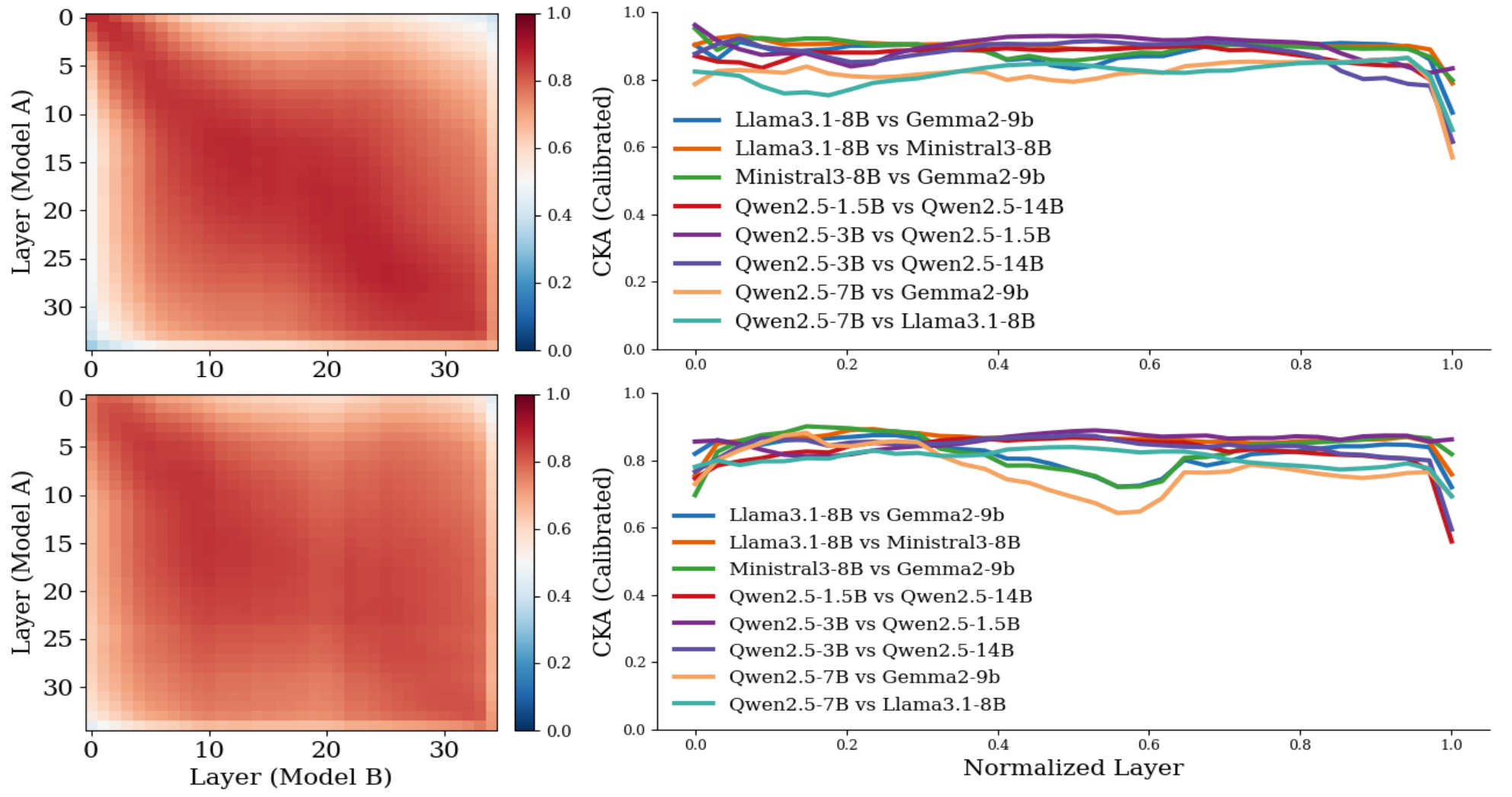} 
  \caption{Cross-model alignment of within and between-concept relations. 
Left: CKA between all layer pairs across two models, with rows indexing layers of model $A$ and columns indexing layers of model $B$. Heatmaps are interpolated to fit the layer differences. Right: diagonal CKA at matched normalized layer depth, shown for eight representative model pairs spanning different families and scales. Upper row: within-concept relation; lower row: between-concept relation.}
  \label{fig:internal}
\end{figure*}

\paragraph{Within and between-concept relations of concepts are aligned across models.} As shown in Figure~\ref{fig:internal}, both reveal a consistent block structure in the layer-pair heatmaps, with high CKA concentrated along and near the diagonal, indicating that representationally similar layers correspond to similar relative depths across architectures and scales. Along the diagonal, CKA remains above 0.7 across the full depth range for most model pairs under both context conditions, with a modest drop near the final layers, and alignment is largely stable across architecturally diverse pairs (e.g., Llama3.1-8B vs.\ Gemma2-9b). These results establish that the representations being transformed are themselves shared across models, a prerequisite for interpreting the transformation field comparisons in the above section.

\paragraph{The alignment cannot be reduced to surface statistics.}
To establish that the shared between-concept relation reflects semantic structure rather than surface distributional properties, we compare LLM-to-LLM alignment under real context (see Section~\ref{subsec:template}) against scrambled context and against static co-occurrence embeddings (Word2Vec, GloVe). The scrambled context pools all words across the six context templates and randomly reassigns them to context positions, preserving token identity and positional information while destroying semantic context structure. Experiment results show that LLM-to-LLM alignment under real context substantially exceeds all other conditions, and the gap widens across layers. Pairwise Mann-Whitney U tests confirm that real-context LLM-to-LLM alignment differs significantly from all other conditions ($p<0.0001$, rank-biserial $r>0.82$). See Appendix~\ref{app:static} for more details on implementation and statistics. Together, these results imply that what models share is not merely distributional habit, but something closer to the structured, context-sensitive organization

\section{Related Work}

\subsection{Concept Representations of Language Models}
Early theoretical work shows that category representations converge to simplex vertices at the final layer of classification networks \citep{papyan2020prevalence}, a phenomenon called neural collapse. \citep{saxe2019mathematical} provides a complementary account showing that semantic concepts decompose into orthogonal bases across properties, which, under classification pressure, produces a similar simplex geometry. \citep{zhaoimplicit, zhao2025geometry} extend both results to LLMs, treating next-token prediction as soft label classification and recovering an analogous geometric structure in the learning dynamics. From a different 
angle, the linear representation hypothesis proposes that concepts correspond to directions in activation space, with semantic hierarchy encoded as orthogonality \citep{park2024linear, parkgeometry}. \citep{huh2024position} extends this to a convergence claim that models approach a shared ideal representation at scale. \citep{engelsnot} complicates this picture, showing that some concepts are inherently multi-dimensional. Across these accounts, 
concepts are treated as static objects. \citep{hu2026representational} challenges this directly, arguing that concepts are constituted by stable relational structure under context, and demonstrating this for number concepts.

\subsection{Context, Concepts, and Neural Geometry}

The idea that context shapes conceptual representation is well-established in both neuroscience and cognitive science. \citep{gao2023context} dissociate 
context-free and context-dependent conceptual representations in distinct cortical networks, while \citep{mante2013context} shows that context routes information through low-dimensional population subspaces in the prefrontal cortex (PFC). \citep{bernardi2020geometry} demonstrates that geometric structure allowing parallel coding directions across conditions in PFC and hippocampus supports cross-condition generalization, paralleling our finding that relational displacement structure transfers across models. At the computational level, \citep{xu2025revealing} shows that few-shot prompting recovers human-like conceptual structure from LLMs. Moreover, \citep{piantadosi2024concepts} argues that vector-based representations provide a compelling account of human concepts, handling similarity, features, and relational structure within a unified format. Our work extends this view dynamically. 

Together, all these works establish that concept representations are stable geometric objects of various kinds, and that contexts act as operators on those representations. But none of the work systematically characterizes the geometry of transformation itself - whether it is structured, semantically organized, or shared across models. Our work attempts to fill this gap by studying the structure of contextual transformation as an inherent property of concepts, and in doing so, offers a geometric account of what it means for a concept to be defined not only by where it is, but by how it moves.

\section{Limitation and Discussion}

Our findings are subject to several boundaries worth noting. First, our evidence comes from autoregressive language models, where context is easily manipulated through prompts. It remains open whether the same geometric structure holds in vision-based models. Second, our stimuli are English only, and the concreteness and density norms we rely on are drawn from English-speaking populations. It thus remains an open question whether the same organization holds across different languages. Third, because we study intermediate rather than final-layer representations, we deliberately avoid analytical claims available under the softmax-calibrated geometry of the unembedding space. Thus, we characterize our geometry non-parametrically rather than deriving a closed-form account, which opens future explorations beyond what we impose here.

The framework we establish in this paper opens new insights for both cognitive science and representational engineering. Classical accounts in cognitive science treat concepts as stable mental representations, prototypes or exemplars \citep{rosch1973internal, nosofsky1986attention}, that are retrieved and subsequently modified by context. Studies from neuroscience have refined this picture, identifying distinct cortical substrates of conceptual representations under different contexts \citep{gao2023context, mante2013context}. Yet a mechanistic account of why concepts can be both stable and flexible has remained out of reach \citep{truman2024flexible, casasanto2015all}. Our results suggest a potential direction to explore: concepts are stable because the relations among their context-dependent representations remain consistent, and they are flexible because the transformations among them carry rich, semantically organized information.

Our findings also invite the machine learning community to reconsider a foundational assumption. A substantial body of work in representation engineering, model steering, and alignment research rests on the premise that concepts correspond to stable directions or vectors that can be reliably manipulated \citep{arditi2024refusal, park2024linear, jhaharnessing}. Our results suggest that what is being manipulated is richer and more structured than this premise acknowledges. This raises a question worth taking seriously: when we steer or align a representation, what exactly are we moving?








%% file: appendix.tex
\label{appendix}

\appendix

\section{Model Details}
\label{app:models}
 
This appendix provides a comprehensive overview of the 23 pre-trained language models
evaluated in this work, spanning six model families released between 2023 and 2025.
Table~\ref{tab:models} summarizes the key statistics of each model;
the following paragraphs describe the architecture and training data for each family.
 
\paragraph{Qwen2.5 (Alibaba / Qwen Team).}
The Qwen2.5 series~\cite{qwen2025qwen25} comprises decoder-only Transformers sharing the same architecture as Qwen2: grouped-query attention (GQA), SwiGLU activations, rotary positional embeddings (RoPE), QKV bias in the attention mechanism, and pre-norm with RMSNorm. The pre-training corpus was scaled from 7 to 18 trillion tokens, with particular focus on knowledge breadth, coding, and mathematics. 
We evaluate six dense open-weight sizes ranging from 0.5B to 32B parameters.
 
\paragraph{Llama (Meta AI).}
We include models from two generations of Meta's Llama series. Llama~2~\cite{touvron2023llama} was trained on approximately 1.8 trillion tokens of publicly available web text, books, and general knowledge data. All dense variants use a standard decoder-only Transformer with SwiGLU activations and RoPE positional embeddings; the 7B and 13B models use standard multi-head attention (MHA), while GQA is only introduced in the 70B model. Llama~3 and Llama~3.1~\cite{grattafiori2024llama} substantially scale the training corpus to 15.6 trillion tokens, with a data mix of roughly 50\% general knowledge, 25\% math and reasoning, 17\% code, and 8\% multilingual content. The architecture adopts GQA with 8 key-value heads across all sizes, increases the RoPE base frequency to 500{,}000, and expands the vocabulary to 128K tokens to better support non-English languages. We evaluate the 8B base models from both Llama~3 and Llama~3.1, as well as the 7B and 13B Llama~2 base models.
 
\paragraph{DeepSeek-LLM (DeepSeek AI).}
DeepSeek-LLM-7B~\cite{bi2024deepseek} follows the same auto-regressive Transformer decoder architecture as LLaMA, with multi-head attention (MHA), SwiGLU activations, RoPE positional embeddings, and RMSNorm. It was trained from scratch on 2 trillion tokens in both English and Chinese, using a multi-step learning rate schedule rather than the cosine schedule common in contemporaneous models. We include this model as a representative early open-weight LLM from a Chinese AI lab, predating the more recent DeepSeek-V2/V3 series.

\paragraph{Mistral / Ministral (Mistral AI).}
We evaluate four models from Mistral AI. The three Ministral~3 models (3B, 8B, and 14B)~\cite{liu2026ministral} are not trained from scratch but derived from Mistral Small~3.1 (24B) via Cascade Distillation, an iterative prune-and-distill procedure that progressively transfers knowledge from the parent model to smaller children. All three share the same architectural foundation: decoder-only Transformer with GQA (32 query heads, 8 KV heads), SwiGLU activations, RoPE positional embeddings, and RMSNorm, with YaRN applied for long-context extension up to 256k tokens. The 3B model additionally uses tied input--output embeddings. We also evaluate Mistral-Small-24B, the parent model of the Ministral~3 family, as a standalone baseline.

\paragraph{Gemma (Google DeepMind).}
We include models from both the Gemma~2~\cite{team2024gemma} generations. Both generations use a decoder-only Transformer with GQA, GeGLU activations, RoPE, and an interleaved local/global attention pattern. Gemma~2 (2B and 9B) additionally applies pre-norm and post-norm with RMSNorm and logit soft-capping for training stability. The 2B and 9B models are trained with knowledge distillation from a larger teacher, on 2T and 8T tokens respectively, drawn primarily from English web documents, code, and mathematics.

\paragraph{Pythia (EleutherAI).}
The Pythia scaling suite~\cite{biderman2023pythia} was designed specifically to facilitate research on learning dynamics and interpretability across model scales. All six models we evaluate (0.4B--12B) share a decoder-only Transformer architecture based on GPT-3, with three key modifications: rotary positional embeddings (RoPE), parallelized attention and feed-forward sub-layers, and untied input/output embeddings. Crucially, every model in the suite was trained on the exact same data in the exact same order: 300 billion tokens from The Pile~\cite{gao2020pile, biderman2022datasheet}, a diverse English-centric corpus comprising web text, books, code, scientific papers (arXiv), and more. This controlled setup makes Pythia uniquely suited for studying the effect of model scale in isolation. Unlike the other families, Pythia was not optimised for downstream benchmark performance, and we therefore do not report MMLU scores for these models.

\begin{table}[h]
\centering
\caption{
  Pre-trained language models used in this work.
  \emph{Layers} = number of transformer layers;
  \emph{MMLU} = 5-shot accuracy (\%); ``---'' = not reported.
}
\label{tab:models}
\renewcommand{\arraystretch}{1.15}
\footnotesize
\begin{tabular}{L{4.4cm} C{1.15cm} C{1.15cm} C{1.15cm} C{1.15cm}}
\toprule
\textbf{Model} & \textbf{Year} & \textbf{Size (B)} & \textbf{Layers} & \textbf{MMLU} \\
\midrule
 
\familyrow{Qwen2.5 \normalfont\textit{(Alibaba / Qwen Team)}}
\rowcolor{rowalt} Qwen2.5-0.5B & 2024 & 0.5  & 24 & 47.5 \\
                  Qwen2.5-1.5B & 2024 & 1.5  & 28 & 60.9 \\
\rowcolor{rowalt} Qwen2.5-3B   & 2024 & 3    & 36 & 65.6 \\
                  Qwen2.5-7B   & 2024 & 7    & 28 & 74.2 \\
\rowcolor{rowalt} Qwen2.5-14B  & 2024 & 14   & 48 & 79.7 \\
                  Qwen2.5-32B  & 2024 & 32   & 64 & 83.3 \\
 
\midrule
\familyrow{Llama \normalfont\textit{(Meta AI)}}
\rowcolor{rowalt} Llama-2-7b-hf   & 2023 & 7  & 32 & 45.3 \\
                  Llama-2-13B-hf  & 2023 & 13 & 40 & 54.8 \\
\rowcolor{rowalt} Llama-3-8B & 2024 & 8  & 32 & 66.7 \\
                  Llama-3.1-8B    & 2024 & 8  & 32 & 66.7 \\
 
\midrule
\familyrow{DeepSeek \normalfont\textit{(DeepSeek AI)}}
\rowcolor{rowalt} deepseek-llm-7b-base & 2023 & 7 & 30 & 48.2 \\
 
\midrule
\familyrow{Mistral / Ministral \normalfont\textit{(Mistral AI)}}
\rowcolor{rowalt} Mistral-Small-24B-Base-2501 & 2025 & 24 & 40 & 80.7 \\
                  Ministral-3-3B-Base-2512    & 2025 & 3  & 26 & 70.7 \\
\rowcolor{rowalt} Ministral-3-8B-Base-2512    & 2025 & 8  & 34 & 76.1 \\
                  Ministral-3-14B-Base-2512   & 2025 & 14 & 40 & 79.4 \\
 
\midrule
\familyrow{Gemma \normalfont\textit{(Google DeepMind)}}
\rowcolor{rowalt} gemma-2-2b        & 2024 & 2  & 26 & 52.2 \\
                  gemma-2-9b        & 2024 & 9  & 42 & 71.3 \\          
                  
\midrule
\familyrow{Pythia \normalfont\textit{(EleutherAI)}}
\rowcolor{rowalt} Pythia-0.4B & 2023 & 0.4 & 24 & --- \\
                  Pythia-1B   & 2023 & 1.0 & 16 & --- \\
\rowcolor{rowalt} Pythia-1.4B & 2023 & 1.4 & 24 & --- \\
                  Pythia-2.8B & 2023 & 2.8 & 32 & --- \\
\rowcolor{rowalt} Pythia-6.9B & 2023 & 6.9 & 32 & --- \\
                  Pythia-12B  & 2023 & 12  & 36 & --- \\

\bottomrule
\end{tabular}
\end{table}


\section{Additional Results for Transformation Field Dispersion}
\label{app:variance}

\begin{figure}[h]
  \centering
  \includegraphics[width=0.99\textwidth]{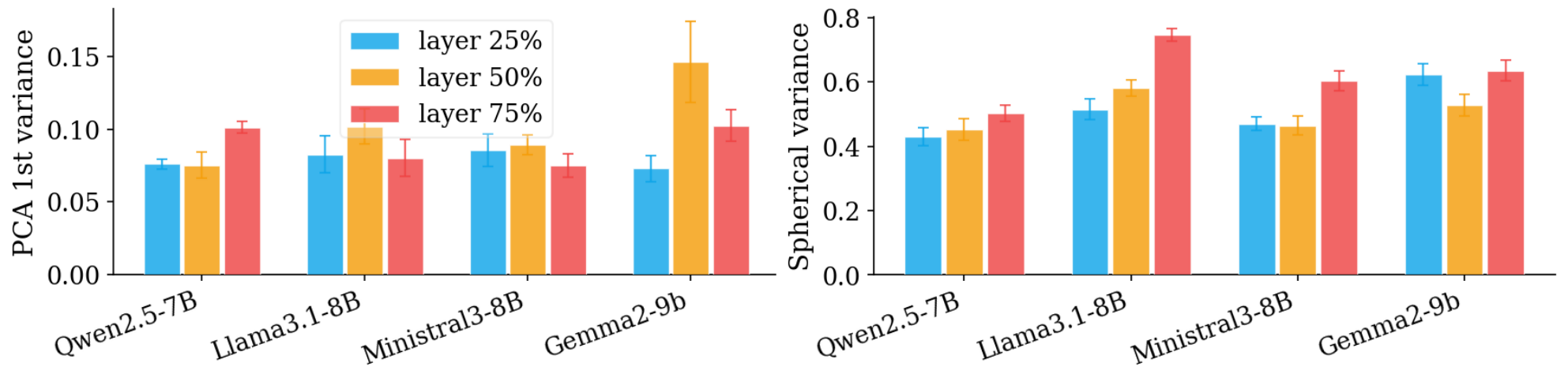} 
  \caption{Transformation field dispersion statistics across four models and three layer depths (25\%, 50\%, 75\%). \texttt{Left:} Leading PCA variance ratio $\rho_1$, defined as the fraction of total displacement variance captured by the first principal component. Values range from 0.073 to 0.146, indicating that no single direction accounts for more than approximately 15\% of the total variance. \texttt{Right:} Global spherical variance $V_\text{global}$, measuring the mean squared angular spread of unit-normalized displacement vectors around their spherical Fr\'{e}chet mean. Values range from 0.43 to 0.75 rad$^2$, reflecting substantial directional heterogeneity across the concept vocabulary. Both statistics are consistent across architecturally diverse models, confirming that contextual displacement does not reduce to a single steering vector at any layer depth.}
  \label{fig:variance}
\end{figure}

Figure~\ref{fig:variance} reports the two field-dispersion statistics, the leading PCA variance ratio $\rho_1$ and the global spherical variance $V_{\text{global}}$.

\paragraph{PCA leading variance ratio.} Across all four models and three layer depths, $\rho_1$ ranges from 0.073 to 0.146, confirming that no single principal direction accounts for more than approximately 15\% of the total displacement variance. The values remain far below what a near-uniform field would require, ruling out a single-vector characterization.

\paragraph{Global spherical variance.} $V_{\text{global}}$ ranges from 0.43 to 0.75 radians$^2$, reflecting substantial directional spread of unit-normalized displacement vectors around their spherical Fréchet mean. The spread is systematically larger at layer depth 75\% except the Gemma2-9b.

Taken together, both statistics converge on the same conclusion: contextual displacement does not reduce to a shared steering direction. The residual variance is large, structured, and consistent across models.


\section{Comparison with Static Embeddings and Scrambled-Context Baseline}
\label{app:static}

We verify that cross-model relational geometry reflects 
semantic structure rather than surface distributional 
properties by comparing LLM-to-LLM alignment under real 
context against two controls. The scrambled-context 
baseline pools all words across the six context templates 
and randomly reassigns them to context positions, 
preserving token identity and positional structure while 
destroying syntactic and semantic context. The static 
embedding controls compare LLM representations against 
Word2Vec and GloVe, treating each as a kernel matrix over 
the shared vocabulary. Since CKA operates on kernel 
matrices and is invariant to embedding dimensionality, 
this comparison is well-defined despite dimensional 
mismatch between LLM hidden states and static embeddings.

\begin{figure}[h]
  \centering
  \includegraphics[width=0.95\textwidth]{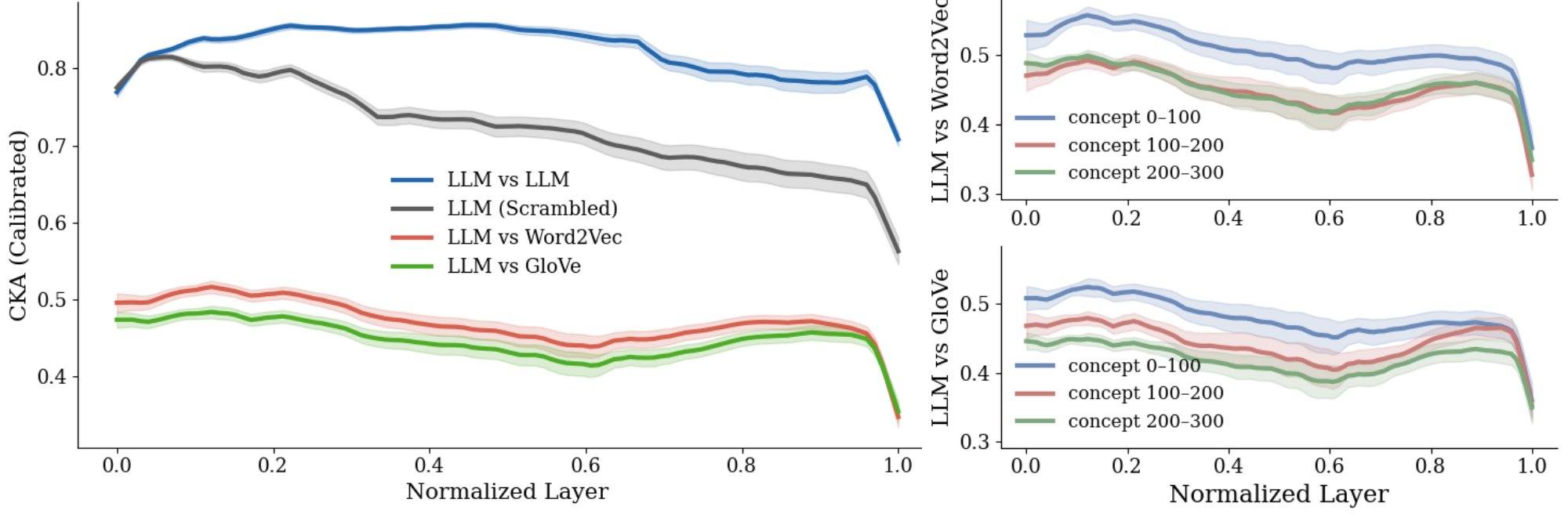} 
  \caption{Relational geometry alignment across conditions. 
Left: CKA between model pairs under real context as defined in Section~\ref{subsec:template} (LLM vs LLM), scrambled context (LLM Scrambled), and against static embeddings (LLM vs Word2Vec, LLM vs GloVe), aggregated across all context types, lexical density subsets, and model 
pairs, plotted against normalized layer depth. Shaded regions 
denote 95\% confidence intervals. Right: LLM-to-static-embedding 
alignment stratified by lexical density subset (concept 0–100: 
highest density; concept 100–200; concept 200–300: lowest 
density), for Word2Vec (top) and GloVe (bottom).}
  \label{fig:static}
\end{figure}

As shown in Figure~\ref{fig:static} (left), LLM-to-LLM alignment under real context substantially exceeds both controls at all layers, with the gap relative to the scrambled baseline widening in deeper layers — consistent with increasing reliance on semantic information at depth. Pairwise Mann-Whitney U tests confirm that real-context LLM-to-LLM alignment is significantly higher than all other conditions ($p < 0.0001$, rank-biserial $r > 0.82$), ruling out surface-level lexical co-occurrence as an explanation.

Kruskal-Wallis tests further reveal that scrambled-context LLM-to-LLM alignment is unaffected by lexical density ($p = 0.7766$), whereas LLM-to-static-embedding alignment varies significantly with lexical density ($p < 0.0001$). As shown in Figure~\ref{fig:static} (right), higher-density concepts (ranked 0–100) yield consistently higher CKA scores against both Word2Vec and GloVe than lower-density subsets, across all layers.


\section{Concept Vocabulary for Internal Geometry Analysis}
\label{app:vocabulary}

The internal geometry analysis requires a concept vocabulary that spans the full range of concreteness while maintaining a balanced artificial vs.\ natural concept distribution. We sample artificial vs.\ natural concepts from WordNet \citep{miller1995wordnet} and define concreteness using Brysbaert's rating \citep{brysbaert2014concreteness}. Note that artificial and natural concepts are inherently concrete concepts. We denote concept above the mid point of rating as concrete concepts (some of them are also artificial and natural concepts) and concept below as abstract concepts as shown in Table~\ref{tab:vocab}. 

\begin{table}[h]
\centering
\caption{Concept vocabulary used in the internal geometry analysis, sorted by concreteness rating.}
\label{tab:vocab}
\small
\begin{tabular}{llc||llc}
\toprule
\textbf{Word} & \textbf{Category} & \textbf{Conc.} & \textbf{Word} & \textbf{Category} & \textbf{Conc.} \\
\midrule
belief      & Abstract & 1.19 & curriculum  & Concrete & 3.23 \\
hope        & Abstract & 1.25 & overlay     & Artifact & 3.30 \\
goodness    & Abstract & 1.38 & instruction & Concrete & 3.37 \\
absolutism  & Abstract & 1.46 & layer       & Artifact & 3.52 \\
idea        & Abstract & 1.61 & decoration  & Artifact & 3.53 \\
wish        & Abstract & 1.77 & mechanism   & Natural  & 3.67 \\
doubt       & Abstract & 1.79 & nursing     & Concrete & 3.76 \\
thought     & Abstract & 1.97 & course      & Concrete & 3.82 \\
culture     & Abstract & 2.04 & universe    & Natural  & 3.85 \\
major       & Abstract & 2.16 & weight      & Concrete & 3.94 \\
conception  & Abstract & 2.24 & core        & Natural  & 3.96 \\
health      & Abstract & 2.28 & fixture     & Artifact & 4.17 \\
density     & Abstract & 2.52 & panel       & Artifact & 4.26 \\
welfare     & Abstract & 2.57 & barrier     & Natural  & 4.41 \\
credit      & Abstract & 2.71 & sill        & Natural  & 4.59 \\
proposal    & Abstract & 2.75 & road        & Artifact & 4.75 \\
factor      & Abstract & 2.79 & canvas      & Artifact & 4.78 \\
series      & Abstract & 2.92 & nest        & Natural  & 4.86 \\
nature      & Natural  & 2.92 & folder      & Artifact & 4.93 \\
user        & Concrete & 3.16 & carpet      & Natural  & 4.96 \\
\bottomrule
\end{tabular}
\end{table}

Table~\ref{tab:vocab} lists all 40 words sorted by concreteness rating. The distribution reflects the intended design: abstract concepts occupy the low end of the scale, while artifact and natural concepts are concentrated in the mid-to-high range, with moderate overlap around the 3.0--4.0 region. This spread ensures that any observed correlation between concreteness and transformation field statistics is not an artifact of a bimodal or truncated distribution.

\section{Leading Variance Ratio $\rho_1$ and Global Spherical Variance}
\label{app:nonlinear}
The leading variance ratio $\rho_1$, defined as the fraction of total variance captured by the leading principal component of the field \citep{abdi2010principal}:
\begin{equation}
    \rho_1 = \frac{\lambda_1(\hat{\Phi}_\tau^{(m)})}{\sum_k \lambda_k(\hat{\Phi}_\tau^{(m)})},
\end{equation}
where $\hat{\Phi}_\tau^{(m)}$ denotes the row-normalized transformation field and $\lambda_k$ its $k$-th eigenvalue of the covariance; $\rho_1$ measures the fraction of total variance captured by the leading principal component of the field. The global spherical variance is defined as:
\begin{equation}
    V_{\text{global}} = \frac{1}{|\mathcal{W}|}\sum_{w \in \mathcal{W}} d_{\mathcal{S}}^2\!\left(\hat{\phi}^{(m)}(w,\tau),\, \mu^*\right),
\end{equation}
where $\hat{\phi}^{(m)}(w,\tau) = \phi^{(m)}(w,\tau)/\|\phi^{(m)}(w,\tau)\|$ is the unit-normalized displacement, $d_{\mathcal{S}}(\mathbf{u}, \mathbf{v}) = \arccos(\mathbf{u}^\top \mathbf{v})$ is the geodesic distance on the unit hypersphere $\mathcal{S}^{d-1}$, and
\begin{equation}
    \mu^* = \argmin_{\mathbf{u} \in \mathcal{S}^{d-1}} \sum_{w \in \mathcal{W}} d_{\mathcal{S}}^2\!\left(\hat{\phi}^{(m)}(w,\tau),\, \mathbf{u}\right)
\end{equation}
is the spherical Fr\'{e}chet mean of the normalized field. $V_{\text{global}}$ thus measures the mean squared angular spread of displacement directions in units of radians$^2$.